\documentclass{article}

\usepackage[preprint]{neurips_ai4d3_2023}

\usepackage[utf8]{inputenc} 
\usepackage[T1]{fontenc}    
\usepackage{hyperref}       
\usepackage{url}            
\usepackage{booktabs}       
\usepackage{amsfonts}       
\usepackage{nicefrac}       
\usepackage{microtype}      
\usepackage{xcolor}         

\usepackage{multirow}
\usepackage{bm}
\usepackage{amsmath}
\usepackage{verbatim}
\usepackage{xspace}

\newcommand{\molga}{\textsc{mol\_ga}\xspace}

\title{Genetic algorithms are strong baselines for molecule generation}

\author{Austin Tripp \\
    University of Cambridge \\
    \texttt{ajt212@cam.ac.uk} \\
    \And
    José Miguel Hernández-Lobato \\
    University of Cambridge \\
    \texttt{jmh233@cam.ac.uk} \\
}

\begin{document}

\maketitle

\begin{abstract}
    Generating molecules, both in a directed and undirected fashion,
    is a huge part of the drug discovery pipeline.
    Genetic algorithms (GAs) generate molecules by
    randomly modifying known molecules.
    In this paper we show that GAs
    are very strong algorithms for such tasks,
    outperforming many complicated machine learning methods:
    a result which  many researchers may find surprising.
    We therefore propose insisting during peer review that new algorithms
    must have some clear advantage over GAs, which we call 
    the \emph{GA criterion}.
    Ultimately our work suggests that a lot of research
    in molecule generation should be re-assessed.
\end{abstract}

\section{Introduction}

Drug discovery is ultimately a molecular generation task,
albeit a very challenging one with many objectives and constraints.
Because of this, many works in machine learning for drug discovery focus on generating molecular graphs\footnote{
This is distinct from generating 3D conformations for a given graph structure.
} \citep{du2022molgensurvey}.
There are many variants of this problem:
for example generating novel molecules for virtual screening
or generating molecules with one or more desirable property values.
Advanced machine learning algorithms have been proposed for all such variants \citep{du2022molgensurvey}.

Yet, for all the purported difficulty of this problem
in some respects generating molecules is surprisingly easy.
The rules governing valid molecules are fairly straightforward
(a series of bond constraints taught to most high-school students)
and can be easily checked using freely available software
such as \texttt{rdkit}.
This means that new molecules can be generated simply by adding, removing, or substituting fragments of known molecules.
When repeated iteratively, this strategy is generally called a \emph{genetic algorithm} (GA),
and is well-studied for molecules \citep{jensen2019graph} and in optimization more generally \citep{holland1992genetic}.

In this work we apply GAs to several molecular generation tasks and find that they perform quite well:
often better than complicated methods based on deep learning, and better than they have been reported previously.
Our results suggest that GAs are underused and underappreciated by the machine learning community.
Given their simplicity and availability, we believe that researchers in molecule generation
should primarily be focusing on methods that \emph{complement} genetic algorithms,
and propose the \emph{GA criterion} to evaluate new methods:
that they should offer some sort of advantage over GAs
(more thoughts in section~\ref{sec:discussion}).

\section{Background on genetic algorithms}

Genetic algorithms (GAs)
is a term that describes a broad class of algorithms which all operate similarly.
GAs generally consist of the following steps:
\begin{enumerate}
    \item \textbf{Start with an initial population} $P$ of molecules.
    \item\label{step:sample} \textbf{Sample a subset} $S\subseteq P$ from the population.
        It could be that $S=P$, or $S$ could be biased to contain ``better'' elements of $P$.
    \item\label{step:gen} \textbf{Use $S$ to generate new molecules $N$.}
        The most common ways of doing this are to randomly modify molecules or randomly combine molecules.
        In GAs these operations are often called \emph{mutation} and \emph{crossover}
        because of their similarity to analogous processes in biology.
    \item\label{step:pop} \textbf{Select a new population $P'$ from $P\cup N$}.
        For example, $P'$ could consist of the ``best'' molecules in $P\cup N$.
    \item Set $P\gets P'$ and go back to step~\ref{step:sample}.
\end{enumerate}
The main factor which differentiates GAs from each other is the generation step (\ref{step:gen}).
Many methods have been proposed:
for example modifying string representations of molecules \citep{nigam2021beyond}
or directly modifying molecular graphs \citep{jensen2019graph}.
However, the sampling and population selection steps are clearly also important:
for example, this could be chosen ensure that the population does not become too homogeneous \citep[p.\@ 21]{srinivas1994genetic}.\footnote{
See specifically the section entitled ``Selection mechanisms and scaling''.
}

\section{Experiments and Results}

In this section we run experiments on several different molecule generation benchmarks.
We use the default GA from the \molga package\footnote{Available at \url{https://pypi.org/project/mol-ga/}} as our genetic algorithm,
which uses the following settings:
\begin{itemize}
    \item Samples (step~\ref{step:sample}) are drawn from the top quantiles of the population (details in Appendix~\ref{appendix:quantile-sampling}).
    \item New molecules are generated (step~\ref{step:gen}) using the mutation and crossover operations from \citet{jensen2019graph},
        based on the implementation in the GuacaMol benchmark \citep{brown2019guacamol}.
    \item The population is selected greedily (step~\ref{step:pop}): i.e.\@ the molecules with the highest scores are added.
\end{itemize}

\subsection{Unconditional molecule generation}

We first examine the task of generating molecules which are distinct from a set of reference molecules:
a task which has been considered in a large number of machine learning papers.
If the set of reference molecules is fairly diverse,
intuitively one would expect random perturbations of the reference molecules to be a strong baseline.
Previous works have used the ZINC~250K dataset \citep{sterling2015zinc} as the reference set
and measured success using the validity,
novelty (i.e.\@ molecules not found in the reference set),
and uniqueness of 10\,000 generated molecules.

Table~\ref{tab:unconditional} shows the performance of a variety of algorithms on this benchmark,
mostly taken from previous papers.\footnote{
The lack of error bars is because most papers do not report them,
with the laudable exceptions of \citet{verma2022modular} and \citet{liu2021graphebm}.
}
At the bottom of the table are results for \molga
(obtained by running steps~\ref{step:sample}--\ref{step:gen} once using the reference set as the population)
and the ``AddCarbon'' method \citep{renz2019failure},
which adds a single carbon atom to random molecules from the reference set.
Metrics were calculated using the \texttt{moses} package \citep{polykovskiy2020molecular}.
Virtually all methods perform similarly, at near 100\% on all metrics.
This shows both that generating molecules is not a particularly hard task,
and that more complex algorithms do not generally perform better than simpler ones.
Something missing from this table is \emph{generation speed}:
e.g.\@ how many molecules can be generated per second.
This would clearly determine the total amount of molecules which could be generated with a given computational budget.
Simple methods like genetic algorithms have a clear advantage here
which is not reflected in this table.

\begin{table}  
  \caption{Unconditional generation results on ZINC~250k dataset.
  Results with `$*$' are taken from \citet[Table 3]{popova2019molecularrnn}, 
  with `$\dagger$' from \citet[Table2]{Shi2020GraphAF:}, 
  with `$\ddagger$' from \citet[Table 1]{luo2021graphdf}, 
  with `$\mathsection$' from \citet[Table 3]{verma2022modular},
  and `$\mathparagraph$' from \citet[Table 5]{liu2021graphebm}. 
  All other results are original.
  }
  \label{tab:unconditional}
  \centering
  \begin{tabular}{llrrr}
    \toprule
    Method    & Paper & Validity & Novelty@10k & Uniqueness    \\
    \midrule
    JT-VAE$^*$ &\citet{jin2018junction}     & 99.8\% & 100\% & 100\%    \\  
    GCPN$^*$ &\citet{you2018graph}     & 100\% & 100\% & 99.97\%    \\
    MolecularRNN$^*$ &\citet{popova2019molecularrnn}     & 100\% & 100\% & 99.89\%    \\
    Graph NVP$^\dagger$       &\citet{madhawa2019graphnvp}          & 100\% & 100\% & 94.80\%    \\ 
    Graph AF$^\dagger$   &\citet{Shi2020GraphAF:}                   & 100\% & 100\% & 99.10\%    \\ 
    MoFlow$^\ddagger$ &\citet{zang2020moflow} & 100\% & 100\% & 99.99\%    \\
    GraphCNF$^\ddagger$ &\citet{lippe2020categorical} & 96.35\% & 99.98\% & 99.98\%    \\
    Graph DF$^\ddagger$ &\citet{luo2021graphdf} & 100\% & 100\% & 99.16\%    \\  
    ModFlow$^\mathsection$ &\citet{verma2022modular} & 98.1\% & 100\% & 99.3\%    \\ 
    GraphEBM$^\mathparagraph$ &\citet{liu2021graphebm} & 99.96\% & 100\% & 98.79\%    \\  
    \midrule
    AddCarbon &\citet{renz2019failure} & 100\% & 99.94\% & 99.86\%    \\  
    \molga & & 99.76\% & 99.94\% & 98.60\%    \\  
    \bottomrule
  \end{tabular}
\end{table}

\subsection{Molecule optimization}

Many papers consider the task of finding molecules which optimize a particular objective function $f:\mathcal{M}\mapsto\mathbb{R}$
(denoting the space of molecules as $\mathcal{M}$).
Since many objectives of practical interest (e.g.\@ binding affinity, toxicity)
can at least somewhat be distilled into scalar functions,
this abstract task is clearly important.
Although the no-free-lunch theorem precludes the development of a \emph{general} algorithm for all functions $f$,
most functions of practical interest have the property
that molecules with similar structures usually have similar activities.
This property can be exploited by algorithms.
In particular, this suggests that producing structurally-similar variants of the best known molecules could be a promising strategy for such problems:
essentially what genetic algorithms do.

We evaluate the performance of \molga on the PMO benchmark \citep{gao2022sample}
of molecule optimization tasks with a limit of 10\,000 evaluations of $f$
(chosen because in practice experiments are expensive so the number of function evaluations is generally limited).
Although several genetic algorithms were tested in this benchmark,
they used large generation sizes ($\approx100$) which limits the number of opportunities for algorithms to improve upon molecules from the previous generation,
effectively limiting the possible improvement over the best starting molecule.
To allow for maximal performance, we set the number of molecules produced in each iteration to $5$,
potentially allowing for $\approx2000$ iterations of improvement.

The results of \molga and the two best methods from \citet{gao2022sample}
are shown in Table~\ref{tab:pmo-benchmark-results}.
Not only does \molga outperform the best GA from \citet{gao2022sample},
it actually outperforms \emph{every} method from \citet{gao2022sample}.
This result is somewhat surprising.
We believe this result is likely an artifact of the tuning of the baselines in PMO,
rather than \molga being an especially good method
(given that \molga is essentially Graph GA).

\begin{table}
  \caption{AUC top-10 scores on PMO benchmark \citep{gao2022sample}.
  Highest numbers (across all methods including those from \citet{gao2022sample} not shown in the table) are in \textbf{bold}.}
  \label{tab:pmo-benchmark-results}
  \centering
  \begin{tabular}{c|rrr}

  \toprule
    Method & REINVENT & Graph GA   & \molga   \\  
    Source & \citet{gao2022sample} & \citet{gao2022sample} & Our experiments    \\
    \midrule
    albuterol\_similarity & 0.882$\pm$ 0.006 & 0.838$\pm$ 0.016 &0.896$\pm$0.035\\
    amlodipine\_mpo & 0.635$\pm$ 0.035 & 0.661$\pm$ 0.020 &\textbf{0.688$\pm$0.039}\\
    celecoxib\_rediscovery & 0.713$\pm$ 0.067 & 0.630$\pm$ 0.097 &0.567$\pm$0.083\\
    deco\_hop & 0.666$\pm$ 0.044 & 0.619$\pm$ 0.004 &0.649$\pm$0.025\\
    drd2 & 0.945$\pm$ 0.007 & 0.964$\pm$ 0.012 &0.936$\pm$0.016\\
    fexofenadine\_mpo & 0.784$\pm$ 0.006 & 0.760$\pm$ 0.011 &\textbf{0.825$\pm$0.019}\\
    gsk3b & 0.865$\pm$ 0.043 & 0.788$\pm$ 0.070 &0.843$\pm$0.039\\
    isomers\_c7h8n2o2 & 0.852$\pm$ 0.036 & 0.862$\pm$ 0.065 &0.878$\pm$0.026\\
    isomers\_c9h10n2o2pf2cl & 0.642$\pm$ 0.054 & 0.719$\pm$ 0.047 &\textbf{0.865$\pm$0.012}\\
    jnk3 & \textbf{0.783$\pm$ 0.023} & 0.553$\pm$ 0.136 &0.702$\pm$0.123\\
    median1 & 0.356$\pm$ 0.009 & 0.294$\pm$ 0.021 &0.257$\pm$0.009\\
    median2 & 0.276$\pm$ 0.008 & 0.273$\pm$ 0.009 &\textbf{0.301$\pm$0.021}\\
    mestranol\_similarity & 0.618$\pm$ 0.048 & 0.579$\pm$ 0.022 &0.591$\pm$0.053\\
    osimertinib\_mpo & 0.837$\pm$ 0.009 & 0.831$\pm$ 0.005 &\textbf{0.844$\pm$0.015}\\
    perindopril\_mpo & 0.537$\pm$ 0.016 & 0.538$\pm$ 0.009 &0.547$\pm$0.022\\
    qed & \textbf{0.941$\pm$ 0.000} & 0.940$\pm$ 0.000 &\textbf{0.941$\pm$0.001}\\
    ranolazine\_mpo & 0.760$\pm$ 0.009 & 0.728$\pm$ 0.012 &\textbf{0.804$\pm$0.011}\\
    scaffold\_hop & \textbf{0.560$\pm$ 0.019} & 0.517$\pm$ 0.007 &0.527$\pm$0.025\\
    sitagliptin\_mpo & 0.021$\pm$ 0.003 & 0.433$\pm$ 0.075 &\textbf{0.582$\pm$0.040}\\
    thiothixene\_rediscovery & 0.534$\pm$ 0.013 & 0.479$\pm$ 0.025 &0.519$\pm$0.041\\
    troglitazone\_rediscovery & \textbf{0.441$\pm$ 0.032} & 0.390$\pm$ 0.016 &0.427$\pm$0.031\\
    valsartan\_smarts & \textbf{0.178$\pm$ 0.358} & 0.000$\pm$ 0.000 &0.000$\pm$0.000\\
    zaleplon\_mpo & 0.358$\pm$ 0.062 & 0.346$\pm$ 0.032 & \textbf{0.519 $\pm$ 0.029} \\
    \midrule
    Sum & 14.196 & 13.751 & \textbf{14.708} \\
    Old Rank & 1 & 2 & N/A  \\
    New Rank & 2 & 3 &1  \\
    \bottomrule
  \end{tabular}
\end{table}

\section{Discussion}\label{sec:discussion}

In this paper we have shown that genetic algorithms are very strong baselines for molecular generation tasks,
performing at least as well as many more complicated methods in the unconditional and
single-objective settings.
Although it is tempting to conclude from this that GAs are incredible high-performing algorithms,
this is actually \emph{not} our opinion:
after all, they are just simple heuristic algorithms.
Rather, we view this as an \emph{indictment} of more modern molecular generation methods.
For the amount of research that has been done in this area, it appears that surprisingly little progress has been made!

We think that the main cause of this is poor empirical practices in machine learning research.
Many experiments are done with an explicit desired outcome, typically that some novel algorithm is the best.
A comprehensive evaluation of baseline methods is usually viewed as cumbersome rather than essential,
especially if these baselines might perform well.
We propose to address this by introducing the \emph{GA criterion}:
a rule of thumb that new methods in molecular generation should offer some advantage over GAs.
This advantage can either be empirical (e.g.\@ outperforming GAs in experiments)
but also conceptual: what limitation of randomly changing known molecules does the proposed method
potentially overcome?
We call upon members of this community to enforce this standard in the peer review process.

Our results also suggest another intriguing possibility:
maybe newer algorithms do not outperform GAs because \emph{they are also just generating variants of known molecules},
just in an indirect way.
Indeed, many newer methods are essentially generative models which are trained on a dataset of known molecules.
It would therefore not be surprising if the novel molecules generated by such methods
are simply variants of their training data.
We think investigating this is an important direction for future work.

\subsubsection*{Acknowledgments}
Austin Tripp acknowledges funding via a C~T Taylor Cambridge International Scholarship and the  Canadian Centennial Scholarship Fund.
José Miguel Hernández-Lobato acknowledges support from a Turing AI Fellowship under grant EP/V023756/1.

\bibliographystyle{apalike}
\bibliography{references}

\appendix
\clearpage
\section{Quantile-based sampling}\label{appendix:quantile-sampling}

\citet{srinivas1994genetic} suggests rank-based sampling to avoid premature convergence on page 21.
The quantile-based sampling in \molga is an attempt to preferentially sample molecules
with high objective function values
while not exclusively sampling them
and depending only on the rank of a given sample.
This is accomplished by the following sampling procedure:
\begin{align*}
    u &\sim\mathcal{U}[-3, 0]  \\
    \epsilon &= 10^u
\end{align*}
A molecule is drawn uniformly from the top $\epsilon$ fraction of the population.
The sample $S$ consists of i.i.d.\@ via the procedure above.

In practice, the implementation is a quasi Monte-Carlo version of the above, wherein a uniform grid of samples $u$ is created deterministically.

\end{document}